# Infrared Imaging Empowered by Artificial Intelligence for Pediatric Skeletal Triage: A Narrative Review and Future Perspectives

**Sajad Amiri[1*], Pardis Afshar[2], Elham Anjomshoa[3]**

*[1] Department of Medicine, Tehran University of Medical Sciences, Tehran, Iran*

*[2] Department of Computer Engineering, Sharif University of Technology*

*[3] Department of Computer Science, Shahid Bahonar University of Kerman, Kerman, Iran*

*[*] Corresponding author: amiri_s@alumnus.tums.ac.ir*

## Graphical Abstract

---

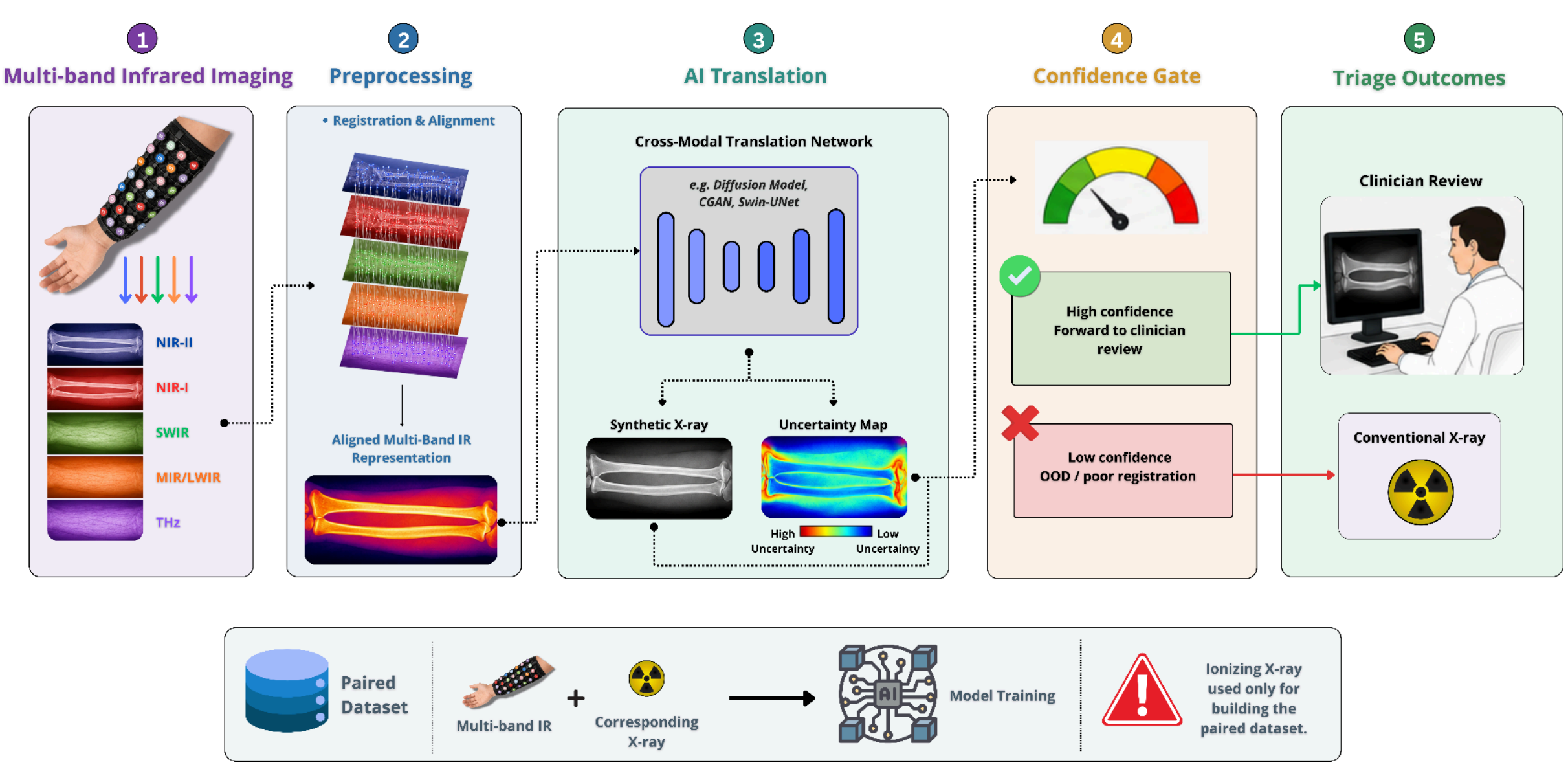

**Graphical Abstract.** Overview of the proposed radiation-free pediatric skeletal-triage framework. Multi-band, non-ionizing infrared data are fused and translated by a deep network into a synthetic radiograph with an associated uncertainty estimate; a confidence gate forwards high-certainty cases to clinician-led triage and escalates low-certainty or out-of-distribution cases to conventional radiography. Ionizing X-ray is used only to build the paired training set.

## Abstract

---

**Background.** Pediatric musculoskeletal trauma represents up to 18% of pediatric ED visits, yet diagnosis still depends on ionizing radiography. Cumulative low-dose radiation in early life raises lifetime leukemia and brain malignancy risk, motivating radiation-free triage alternatives.

**Objective.** To synthesize evidence for a hybrid framework coupling broad-spectrum infrared (IR) imaging with deep-learning cross-modal translation to generate clinically interpretable synthetic-radiograph reconstructions from non-ionizing data.

**Approach.** We review five IR spectral windows spanning 650 nm to 1 mm - NIR-I, NIR-II, SWIR, MIR/LWIR, and THz - and how dual-geometry (transmission/reflection) acquisition exploits wavelength-specific tissue depth and biochemical sensitivity. We summarize image-to-image translation networks (Pix2Pix, CycleGAN, Swin-Unet) and feature-matching algorithms (SuperPoint, SuperGlue, ALIKED, LightGlue) used to align and fuse IR data into radiograph-equivalent reconstructions.

**Implications.** Pediatric anatomy - smaller cross-sections, thinner cortical bone - favors IR penetration, enabling compact, portable, non-ionizing triage hardware. Feasibility is grounded in fNIRS and transcranial photobiomodulation evidence: near-infrared light passes through skin, skull, and cortex with sufficient signal for hemodynamic monitoring - a longer, more attenuating path than through a pediatric forearm or distal leg. Key barriers: paired IR/X-ray dataset construction, AI-as-medical-device regulatory pathways, generalization across body habitus and skin pigmentation, and acquisition-protocol standardization.

**Conclusions.** Integrated multi-spectral IR+AI imaging is a promising radiation-free complement to pediatric skeletal radiography. Progress requires multi-center paired datasets, externally validated models, and IR source safety qualification under IEC 60825-1.

# 1. Introduction

---

Musculoskeletal complaints account for an estimated 8–18% of pediatric emergency department (ED) visits in high-income settings, with trauma — most often contusions, sprains, and distal-extremity fractures — representing the dominant etiology [1, 2, 3]. Distal forearm fractures alone account for roughly 29% of pediatric fractures [4]. The standard diagnostic pathway for suspected pediatric fracture remains plain radiography, with computed tomography (CT) reserved for complex injuries; both modalities expose developing tissues to ionizing radiation.

Children are disproportionately vulnerable to radiation-induced malignancy because of their longer post-exposure life expectancy and the higher proliferation rates of pediatric stem-cell populations. A landmark Australian linkage study of more than 680,000 individuals exposed to CT before age 20 reported a 24% relative increase in cancer incidence per scan [5], paralleling earlier U.S. cohort data linking pediatric CT exposure to leukemia and brain tumors [6]. Although per-examination doses for digital radiography are small, the cumulative dose across years of orthopaedic follow-up, combined with the conservative bias toward repeat imaging in pediatric ED settings, makes dose reduction a public-health priority.

Existing non-ionizing alternatives have important limitations. Magnetic resonance imaging (MRI) provides excellent soft-tissue and marrow contrast but is costly, frequently requires sedation in young children, and is rarely available at the point of care [7, 8]. Point-of-care ultrasound (POCUS) has gained considerable supporting evidence — pooled sensitivity of approximately 90% and specificity of 95% for distal forearm and skull fractures [9, 10, 11] — but performance depends critically on operator training, beam angulation, and acoustic shadowing artifacts that can obscure deeper or comminuted lesions.

A promising third pathway exploits the optical and biochemical heterogeneity of the infrared (IR) spectrum. Recent advances in broad-band detectors, semiconductor laser/LED sources, and computational image translation suggest that paired multi-spectral IR acquisition, processed by deep neural networks, could yield radiograph-equivalent reconstructions suitable for triage of low- to moderate-acuity pediatric skeletal injury [12, 13, 14]. Optical imaging has in fact been reviewed specifically for pediatric disease, where its low cost and absence of ionizing radiation are especially advantageous, even though clinical adoption beyond optical coherence tomography has so far remained limited [49]. The overall concept is summarized in the Graphical Abstract.

### 1.1 Aim and search strategy

This narrative review consolidates evidence across three domains: (i) the biophysics of IR–tissue interaction across the 650 nm – 1 mm range; (ii) deep-learning architectures for cross-modal medical image translation; and (iii) multi-spectral feature matching for image fusion. We searched PubMed, Scopus, IEEE Xplore, and arXiv (January 2000 – April 2026) using combinations of the terms *near-infrared*, *SWIR*, *terahertz imaging*, *diffuse optical*, *functional near-infrared spectroscopy*, *transcranial near-infrared*, *image-to-image translation*, *GAN*, *feature matching*, and *pediatric fracture*. Reference lists of pivotal papers were screened manually. Studies were prioritized by relevance to pediatric applications, clinical translatability, and methodological rigor.

## 2. Broad Infrared Spectrum Imaging Capabilities

---

### 2.1 Photon transport in tissue

The depth at which a useful diagnostic signal can be recovered from biological tissue is governed by the balance between absorption ($\mu_a$) and reduced scattering ($\mu_s'$). In the diffuse regime, this balance is captured by the effective attenuation coefficient:

$$\mu_{eff} = \sqrt{[\, 3\mu_a(\mu_a + \mu_s') \,]}$$

where $1/\mu_{eff}$ approximates the penetration depth of diffuse photons [15]. Within the so-called biological transparency window between approximately 650 nm and 1,350 nm, hemoglobin absorption falls sharply and water absorption remains modest, yielding $\mu_{eff}$ values low enough to permit centimeter-scale transmission through pediatric limbs. This range overlaps the classical phototherapeutic window (approximately 650–1,000 nm); at longer wavelengths water absorption rises steeply, which both limits penetration and can deposit heat in tissue [50]. Recovering the underlying absorption ($\mu_a$) and reduced-scattering ($\mu_s'$) coefficients from measured reflectance and transmittance is itself an active field — steady-state, time-domain, frequency-domain, and spatial-frequency-domain methods each trade measurement accuracy against instrument complexity — and the resulting optical-property maps supply quantitative priors for the downstream synthesis network [51]. The wavelength dependence of this penetration across the infrared spectrum is summarized in Figure 1.

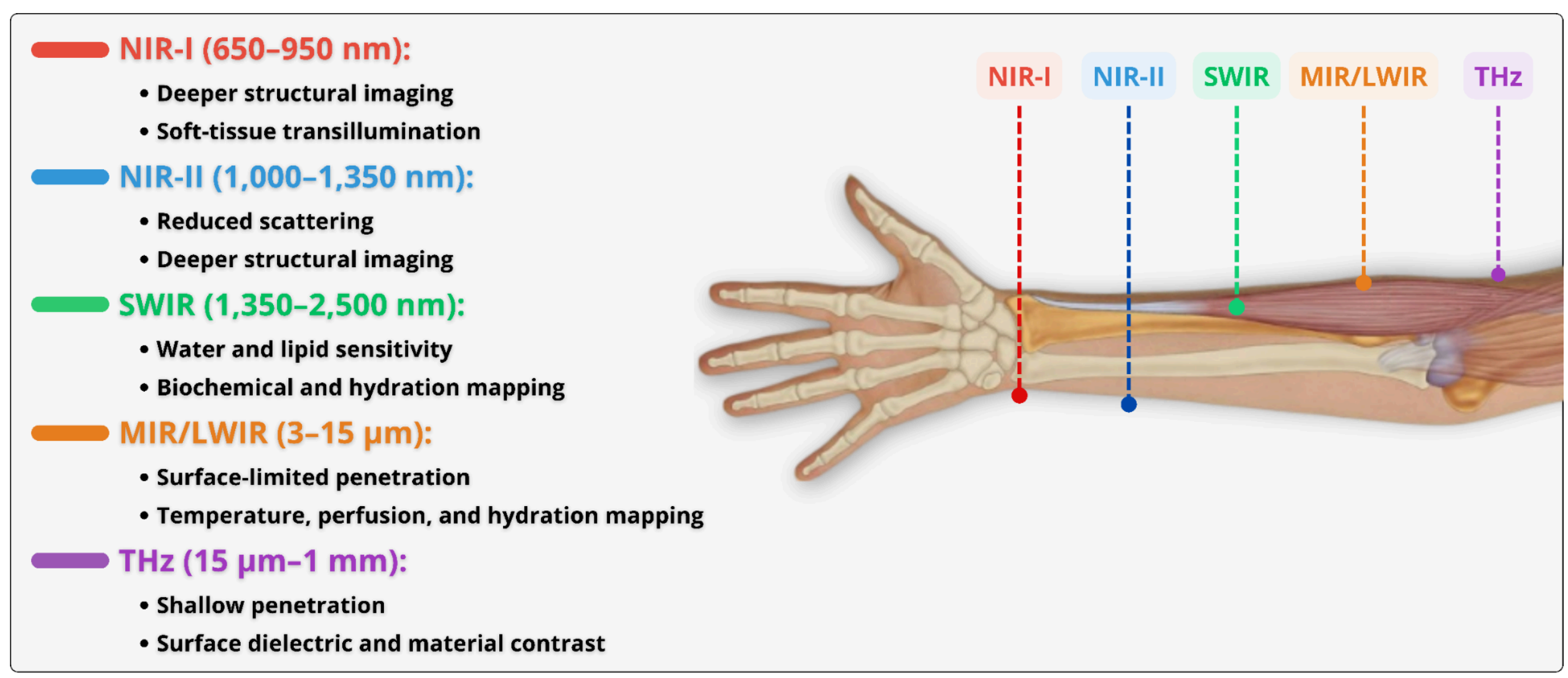

**Figure 1.** Wavelength-dependent penetration of infrared bands through limb tissue. Peak penetration lies in the NIR/SWIR "optical window" (approximately 650–1,350 nm) and falls sharply at longer wavelengths; strong water absorption confines MIR/LWIR and THz to near-surface depths in vivo.

We propose a dual-geometry acquisition scheme (Figure 2): **Transmission (transillumination)** captures deep structural features — including cortical-bone shadowing and gross trabecular geometry — at NIR-I and NIR-II wavelengths; **reflection (backscatter / diffuse reflectance)** maps superficial biochemical content — water, lipid, collagen, and hydration state — at SWIR and MIR wavelengths. Joint analysis allows separation of cortical bone, soft tissue, water and lipid pools, and surface hydration, which together provide structural and chemical priors for the downstream radiograph-synthesis network [12, 16, 17].

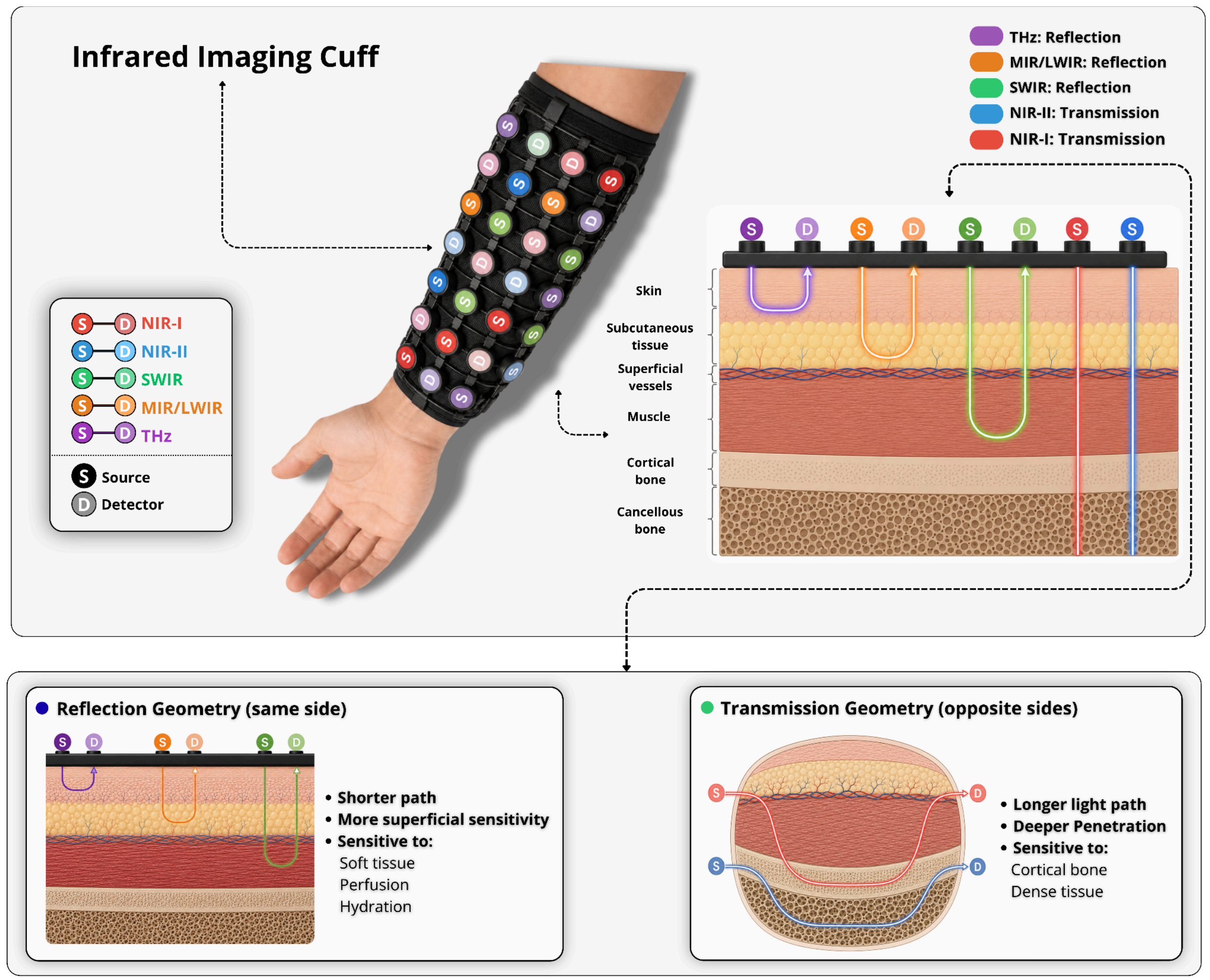

**Figure 2.** Dual-geometry acquisition. In transmission the source and detector sit on opposite sides of the limb and capture deep structural cues (cortical bone and gross morphology) at NIR-I/NIR-II; in reflection they sit on the same side and map superficial biochemical content (water, lipid, hydration) at SWIR and MIR/LWIR. Fusing both geometries combines deep structural and superficial biochemical maps of the same limb.

## 2.2 Direct physiological evidence: near-infrared light already traverses skin, bone, and brain

The single most compelling argument for the feasibility of trans-limb IR imaging is that the harder problem has already been solved in the clinic. Functional near-infrared spectroscopy (fNIRS) demonstrates, every day, that near-infrared light traverses skin, bone, and central nervous tissue in living humans and returns a usable signal. In an adult, NIR light at 700–1,000 nm emitted from a source optode passes through the scalp (3–4 mm), the cranial vault (roughly 5–10 mm of dense cortical bone), the cerebrospinal-fluid layer, and the superficial cortex; a measurable fraction is back-scattered to a detector optode placed 2–3 cm away on the scalp, where its attenuation encodes regional changes in oxygenated and deoxygenated hemoglobin through the modified Beer–Lambert law [55, 56]. The first demonstration that NIR light could cross the intact head and report tissue oxygenation dates to Jöbsis in 1977 [55]; in the decades since, this "banana-shaped" photon path has been refined into a routine bedside tool that maps task-evoked activation in the motor, prefrontal, somatosensory, and language cortices, dissociates ipsilesional from contralesional reorganization after stroke, and tracks neuroplasticity longitudinally with a spatial resolution of roughly 2–3 cm [57, 58, 59].

The same light–tissue physics underlies transcranial photobiomodulation. Reviews of the field report explicitly that emitted NIR light "can penetrate transcranially the physical barriers of the skin and skull and reach the brain parenchyma" when an optical window of approximately 650–1,200 nm is used, without excessive heat generation, and that longer NIR wavelengths scatter less and therefore reach greater depth [60]. Independent groups have imaged neuronal structure and function through the fully intact mouse skull using three-photon microscopy [61], and quantitative red/NIR light dosimetry has been mapped within the human deep brain [62]. Taken together, these literatures establish beyond reasonable doubt that NIR photons cross both soft tissue and dense bone and return information-bearing signal.

The implication for pediatric skeletal triage is direct and, if anything, conservative. If NIR light can complete a round trip through the adult skull — among the densest, most attenuating bony barriers in the body — and still yield spatially resolved, clinically actionable signal, then transillumination of the comparatively thin cortical bone and smaller soft-tissue volume of a pediatric forearm, wrist, or distal lower leg sits well within the demonstrated penetration envelope. The proposed framework therefore introduces no new physics; it repurposes a light–tissue interaction already validated across two decades of neuromonitoring, cerebral and somatic oximetry, and photobiomodulation practice, and redirects it from functional (hemodynamic) sensing toward structural (skeletal) reconstruction.

### 2.3 Wavelength-specific diagnostic content

In the NIR-I window (650–950 nm), low hemoglobin and water absorption allow centimeter-scale soft-tissue transillumination, supporting volumetric mapping of muscle, vasculature, and subcutaneous structures [13, 18].

The NIR-II window (1,000–1,350 nm) — which overlaps with the short end of the SWIR band — combines moderate water absorption with reduced scattering, yielding sharper images of cortical bone and improved structural resolution at the modest depths characteristic of pediatric limbs [12, 19]. Because scattering falls monotonically with wavelength, NIR-II and SWIR light penetrate cleaner and deeper per unit attenuation than NIR-I, a property already exploited to push optical imaging beyond the depths accessible to shorter wavelengths [60, 61].

SWIR imaging (~1,350–2,500 nm) is uniquely sensitive to water and lipid distributions. Zhao et al. demonstrated label-free SWIR meso-patterned imaging that quantifies tissue water and lipid content with high specificity [14], while Mironov et al. used SWIR reflection to discriminate superficial from deep burns [20] — an indirect demonstration that SWIR can localize tissue interfaces.

Mid- and long-wave infrared (MIR/LWIR, 3–15 μm) wavelengths penetrate only superficially because of strong water absorption, but they offer unrivalled sensitivity to surface temperature gradients, perfusion changes, and skin hydration — properties useful for boundary localization and adjunct inflammation mapping [13, 18, 21].

Terahertz imaging (15 μm – 1 mm; ~0.1–10 THz) has shown promise for cortical-bone characterization. Stringer et al. used terahertz time-domain spectroscopy on human cortical bone and detected density-dependent dielectric variations relevant to bone-quality assessment [22]. Although THz penetration in vivo is shallow (millimeters), it is a powerful adjunct modality at the cortex.

That mineralized tissue carries learnable optical signatures is further supported by near-infrared spectroscopy of hydroxyapatite-rich hard tissue: in intact teeth, NIR spectra combined with machine-learning classifiers (support-vector machines, partial-least-squares discriminant analysis, and artificial neural networks) discriminate fine compositional and structural differences with accuracies up to ~95% [63]. Because cortical bone shares the same dominant hydroxyapatite chemistry, this is an encouraging analog for learning skeletal features directly from multi-band IR data.

### 2.4 Hardware feasibility and safety

Compact, low-cost InGaAs sensors for SWIR, bolometers for LWIR, and pulsed photoconductive antennas for THz are now widely commercially available, making multi-band acquisition mechanically and economically feasible at the bedside [23, 24]. Laser- or LED-based illumination must conform to international ocular and dermal exposure limits — most notably IEC 60825-1 [25] — which specify maximum permissible exposures (MPEs) as a function of wavelength, pulse duration, and source geometry. Devices operating below the Class 1 / Class 1C accessible emission limits can be considered safe for routine non-contact pediatric use, an important regulatory advantage over ionizing modalities. Clinically, near-infrared spectroscopy is already a mature bedside technology: continuous-wave, spatially-resolved, time-domain, frequency-domain, and functional NIRS systems quantify tissue oxygenation through the modified Beer–Lambert law, and commercial cerebral and somatic oximeters are in routine perioperative use [52]. This established clinical footprint — alongside two decades of fNIRS deployment in stroke, neonatal, and neuro-critical care [57, 58] — supports the hardware feasibility of compact, multi-band IR acquisition at the point of care.

**Table 1. Infrared spectral windows and their diagnostic role in the proposed AI-fusion framework.**

| Window | Wavelength | Dominant interaction | Diagnostic role | Key refs |
|---|---|---|---|---|
| NIR-I | 650–950 nm | Low $\mu_a$, moderate $\mu_s'$ | Deep soft-tissue transillumination; transcranial & trans-limb optical access (fNIRS-validated) | [13, 18, 55–59] |
| NIR-II | 1,000–1,350 nm | Reduced scattering | Cortical-bone structural mapping | [12, 19] |
| SWIR | 1,350–2,500 nm | Strong water / lipid bands | Biochemical mapping; tissue moisture and burn discrimination | [14, 20] |
| MIR / LWIR | 3–15 µm | Surface water absorption | Boundary localization; perfusion and hydration | [13, 18, 21] |
| THz | 15 µm – 1 mm | Bone dielectric contrast | Cortical-bone density and quality | [22] |

## 3. Artificial Intelligence: Cross-Modal Translation and Multi-Spectral Matching

Converting multi-spectral IR acquisitions into clinically interpretable, radiograph-equivalent images is a cross-modal image-to-image (I2I) translation problem. Three families of methods are most relevant.

### 3.1 Generative adversarial networks for I2I translation

Conditional generative adversarial networks (cGANs), introduced as Pix2Pix by Isola et al. [26], remain the canonical supervised baseline for paired I2I problems. CycleGAN [27] relaxes the paired-data requirement through a cycle-consistency loss, which is attractive in early-stage development where paired IR/X-ray datasets are scarce. Both architectures have been widely repurposed for medical applications, including CT-to-MR translation, low-dose-to-standard-dose synthesis, and modality augmentation [28, 29]. For pediatric skeletal applications, cGANs with anatomically aware loss terms — bone-mask losses and perceptual losses anchored on skeletal atlases — reduce hallucination and improve fidelity at the bone–soft-tissue interface [30]. More recently, denoising diffusion probabilistic models have emerged as a third architecture family for cross-modal medical image synthesis, often achieving superior sample fidelity and calibrated uncertainty estimates compared with adversarial approaches, at the cost of slower inference.

### 3.2 Transformer architectures for global anatomical consistency

Convolutional generators excel at local texture but can drift on global geometry. Vision transformers preserve long-range spatial dependencies and have improved structural fidelity in medical translation tasks. Swin-Unet [31], built on the Swin Transformer's shifted-window self-attention [32], offers an efficient hierarchical encoder–decoder that preserves the global skeletal layout while resolving fine cortical detail — a property particularly valuable for pediatric anatomy, in which growth plates and ossification centers vary substantially with age.

### 3.3 Multi-spectral image matching and fusion

Because each IR band captures complementary information at potentially different exposures and slightly different geometries, robust correspondence matching is a prerequisite for fusion. Modern deep local features — SuperPoint [33], SuperGlue [34], ALIKED [35], and LightGlue [36] — provide rotation- and illumination-invariant keypoint detectors and graph-neural-network matchers that have superseded classical descriptors (SIFT, ORB) under radiometric variation. The Deep-Image-Matching toolbox [37] consolidates these into a reproducible pipeline applicable to multi-view, multi-spectral acquisition. For pediatric IR imaging, these matchers enable subpixel alignment of NIR, SWIR, and LWIR frames before downstream translation, and are robust to subtle pediatric motion and the radiometric differences between bands.

### 3.4 Combined pipeline

The complete pipeline therefore comprises: (a) multi-band IR acquisition in transmission and reflection geometries; (b) deep-feature matching for per-pixel alignment; (c) a cross-modal generator (cGAN or Swin-Unet) trained on paired IR/X-ray data; and (d) anatomical-consistency post-processing with uncertainty estimation [38]. Figure 3 illustrates this pipeline end to end, including the confidence gate that routes low-certainty reconstructions to conventional imaging rather than returning them as diagnostic output.

**AI pipeline**

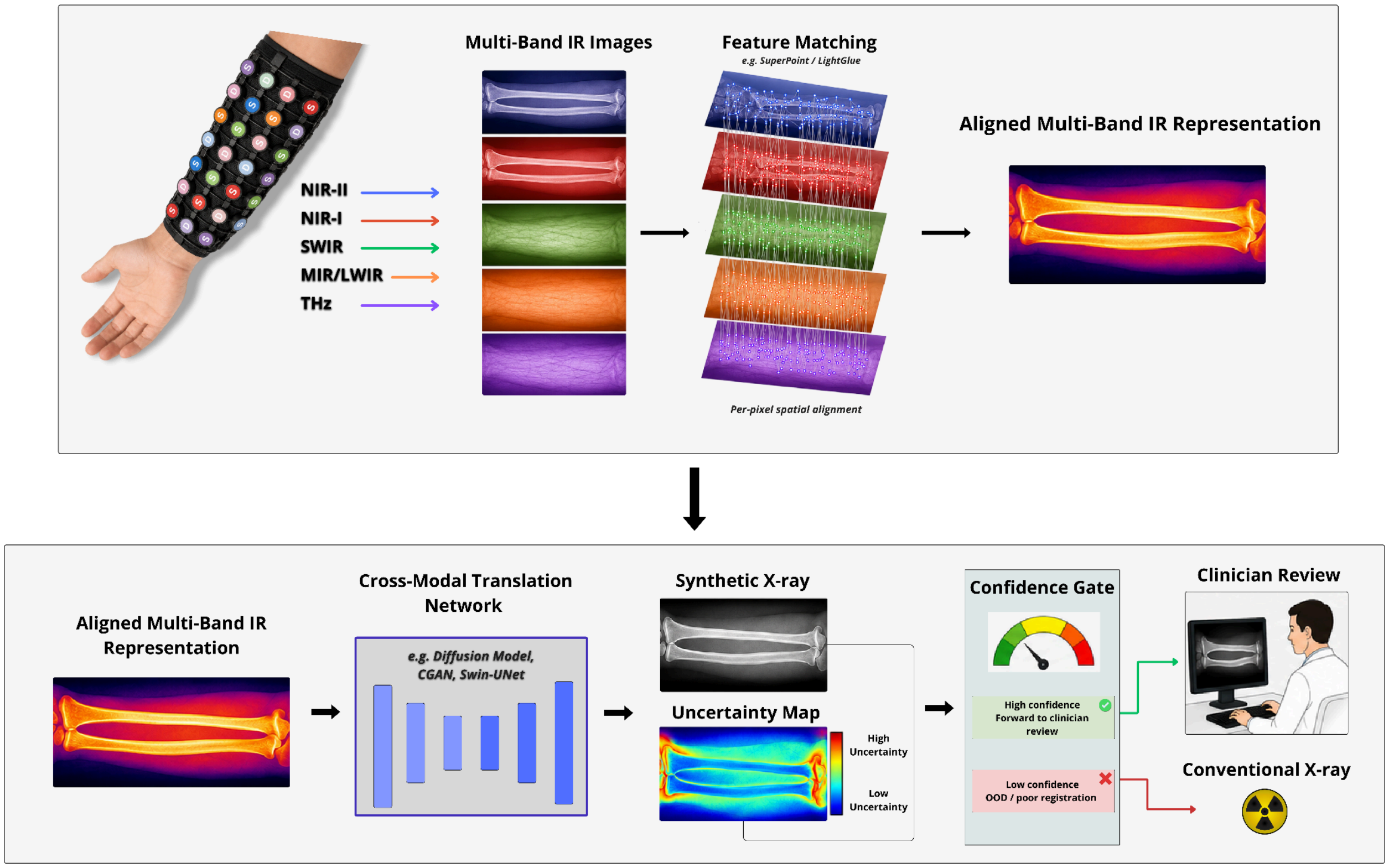

**Figure 3.** AI pipeline for multi-spectral matching, fusion, translation, and confidence-gated output. Multi-band IR images are registered and matched (e.g. SuperPoint, LightGlue), fused, and passed to a cross-modal translation network (e.g. Diffusion, Swin-UNet) regularized by a skeletal-atlas anatomical-consistency loss. The network emits a synthetic radiograph and an uncertainty map; a confidence gate forwards high-certainty cases to clinician review and escalates low-confidence or out-of-distribution cases to conventional X-ray.

## 4. Future Perspectives

---

### 4.1 Why pediatrics first

Pediatric populations are an unusually favorable initial target. Pediatric limbs are thinner and contain proportionally less adipose tissue, which improves NIR/SWIR penetration and signal-to-noise ratio. Bone cortices are thinner; growth plates produce characteristic radiographic features that may be more recoverable from optical data; and clinical practice already accepts a high proportion of “rule-out fracture” examinations that ultimately do not require ionizing imaging [4, 39]. Even a modest specificity gain at the triage stage could substantially reduce radiograph utilization.

This penetration advantage is even more pronounced in early life: neonatal and infant skulls are markedly thinner and only partially ossified, with open fontanels still present in the first year, and fNIRS is already an established bedside technique for non-invasive cerebral oxygenation monitoring in newborns, preterm infants, and infants undergoing cardiac surgery [56, 57, 58]. The same tissue properties that render the pediatric cranium relatively transparent to NIR also render pediatric limbs optically favorable — markedly more so than the adult skull that fNIRS already penetrates routinely — reinforcing pediatrics as the natural first translational target for an IR + AI skeletal-triage framework.

### 4.2 Paired-dataset construction

The single greatest obstacle to clinical translation is the construction of paired IR/X-ray datasets. We propose that, when X-ray is clinically indicated and ethically obtained, simultaneous or near-simultaneous IR acquisition under IRB- and parental-consent-approved protocols could yield curated paired data without any incremental ionizing exposure [40]. Federated multi-center pipelines that keep data within hospital networks would mitigate privacy concerns while expanding age, body-habitus, and skin-tone diversity, which remain known sources of optical-imaging bias [41, 42]. During training, spatially registered X-ray images are used as labels and paired IR images are used as inputs to fit the cross-modal model; ionizing X-ray is therefore required only for constructing the dataset. At deployment, the device acquires IR data alone, and the trained model generates a synthetic radiograph together with an uncertainty estimate. High-certainty cases proceed to clinician triage, whereas low-confidence cases are escalated to conventional X-ray.

### 4.3 Hardware and deployment

The hardware footprint required is modest: a multi-LED/laser-diode source, an InGaAs or pyroelectric detector array, a single-board computer for inference, and a tablet display. Such systems are deployable on ambulances, mobile pediatric units, school clinics, and humanitarian settings where conventional radiography is unavailable [23, 43]. The non-ionizing nature of the modality eliminates lead-shielding and dosimetry requirements, simplifying logistics considerably. Hardware miniaturisation here can borrow directly from the wearable fNIRS community, where multi-channel optode caps with on-board inference now operate at the bedside and in ambulatory settings [57, 59].

### 4.4 Regulatory roadmap

Clinical deployment will follow the standard AI-as-a-medical-device (SaMD) regulatory pathway. Key milestones include device safety qualification under IEC 60825-1 [25] and IEC 60601-1 [44]; analytical and clinical validation aligned with FDA and EMA guidance on adaptive AI [45]; and post-market performance monitoring to detect distribution shift, particularly across pediatric BMI strata and skin-pigmentation phenotypes [42]. Transparent reporting of model development and evaluation — for example following the Checklist for Artificial Intelligence in Medical Imaging (CLAIM) — will be essential for peer review, reproducibility, and ultimately regulatory acceptance [53].

## 5. Discussion

---

### 5.1 Strengths of the framework

The proposed framework offers four distinctive advantages: (i) non-ionizing safety, with cumulative-dose risk effectively zero under IEC 60825-1 Class 1 operation; (ii) biochemical specificity beyond what either X-ray or ultrasound provides — water, lipid, hydration, and (with THz) bone-density information are accessible simultaneously; (iii) portability suitable for point-of-care and pre-hospital triage; and (iv) scalability through standard digital cameras and embedded inference, rather than capital-intensive infrastructure. Critically, the optical feasibility of the approach rests not on speculative physics but on a large, mature clinical evidence base: NIR light is already used to probe oxygenation and to deliver photobiomodulation through skin, skull, and cortex in adults, neonates, and stroke survivors [55–62], and the pediatric limb is a categorically easier optical target than the structures these modalities already interrogate routinely.

### 5.2 Limitations and open challenges

Several limitations must be acknowledged. Depth penetration of longer-wavelength bands (MIR/LWIR, THz) drops sharply in patients with higher BMI or larger limb diameters; in such cases the NIR-II and SWIR-I bands will dominate the diagnostic signal [16, 22]. Model generalizability is constrained by the size and diversity of available training data — historical optical-imaging studies have systematically under-sampled darker skin pigmentations and obesity, and AI radiology models exhibit performance degradation under distribution shift [41, 46]. A further safety-critical concern is specific to generative translation: cross-modal networks can synthesize anatomically plausible but spurious detail (hallucination), so every reconstruction must carry calibrated pixel- and image-level uncertainty, and a mandatory escalation pathway must route low-confidence or out-of-distribution cases to conventional radiography rather than return a potentially misleading image (Graphical Abstract; Figures 3 and 4) [38]. Clinical workflow integration in high-throughput EDs will require sub-30-second acquisition and inference latency, robust calibration, and an interpretable output that emergency clinicians can act on without specialist training [47]. Ethical and logistical considerations around paired-data acquisition are non-trivial: pediatric consent, data governance, and the duty not to delay clinically indicated radiography for research purposes must all be respected [48].

Several practical limitations articulated for adult fNIRS — penetration-depth shortening in patients with thick scalp and skull, sensitivity to motion artefacts, and physiological drift from cardiac and respiratory cycles — converge with the limitations facing any optical modality acquired through dense tissue [57]. Encouragingly, the mitigation strategies developed by the fNIRS and diffuse-optics communities over the past two decades transfer directly to multi-band IR limb imaging: short-separation reference channels to subtract extracerebral / extra-skeletal noise; multi-distance source–detector layouts that disentangle superficial from deep contributions; tightly clocked illumination paired with motion-tolerant deep-learning denoising; selection of longer, less-scattered NIR-II wavelengths to recover depth [60, 61]; and multimodal fusion with EMG or accelerometry for artefact rejection [57, 59]. Importing this engineering vocabulary lowers the technological risk of clinical translation considerably.

### 5.3 Comparison with point-of-care ultrasound

It is essential to position this framework against POCUS, which is already partially displacing radiography for selected pediatric fracture indications and is itself non-ionizing [9, 11]. POCUS has the advantages of mature training infrastructure and FDA-cleared devices. The proposed IR + AI framework would complement, not replace, POCUS: it requires no operator skill at an acoustic interface, scans whole-limb volumes in a single capture, and provides biochemical contrast that ultrasound cannot. Beyond ultrasound, emerging optoacoustic (photoacoustic) imaging — which detects ultrasound generated by absorbed near-infrared light and reaches millimeter- to centimeter-scale depths at sub-millimeter resolution — is a further non-ionizing modality that the framework could complement rather than displace [54]. A reasonable clinical evolution would deploy IR + AI as automated first-pass triage, with POCUS or radiography reserved for cases flagged as positive or uncertain.

## 6. Conclusion

---

Multi-spectral infrared imaging, paired with modern deep-learning translation and feature-matching architectures, offers a credible path toward radiation-free pediatric skeletal triage. Wavelength-selective acquisition exploits the depth- and biochemistry-specific properties of the broad IR spectrum, and AI-driven cross-modal translation can render the fused result in a clinically familiar radiograph-equivalent format. Pediatric anatomy is uniquely suited as a first translational target, and the optical-physics feasibility argument is no longer speculative: decades of fNIRS and transcranial photobiomodulation practice already demonstrate that NIR light traverses skin, skull, and cortex with diagnostically actionable signal recovery in adults, neonates, and clinical neurorehabilitation cohorts [55–62], so penetrating the much thinner intervening tissue of a pediatric forearm or distal lower leg is technologically modest by comparison. The principal next steps are the construction of paired multi-center IR/X-ray datasets, externally validated translation models, hardware qualification under IEC 60825-1, and prospective clinical comparison with both radiography and point-of-care ultrasound.

## Declarations

---

**Funding.** No specific funding was received for this work.
**Conflicts of interest.** The authors declare no competing interests.
**Data availability.** Not applicable; this is a narrative review and no new data were generated.

## References

---

[1] M. P. Mintegi-Raso et al., “Epidemiology of musculoskeletal pain in a pediatric emergency department,” Rheumatology International, vol. 35, no. 11, pp. 1791–1797, 2015, doi: 10.1007/s00296-015-3335-9.

[2] R. Kraus and K. Dresing, “Rational usage of fracture imaging in children and adolescents,” Diagnostics (Basel), vol. 13, no. 3, p. 538, 2023, doi: 10.3390/diagnostics13030538.

[3] P. Langevin et al., “Direct-access physiotherapy to improve access to quality care for children and adolescents presenting to the pediatric emergency department with musculoskeletal problems: the PEDPTMSK pilot randomized control trial,” Journal of Orthopaedic and Sports Physical Therapy, vol. 55, pp. 440–450, 2025, doi: 10.2519/jospt.2025.13321.

[4] R. M. Hedström et al., “Epidemiology of fractures in children and adolescents,” Acta Orthopaedica, vol. 81, no. 1, pp. 148–153, 2010, doi: 10.3109/17453671003628780.

[5] J. D. Mathews et al., “Cancer risk in 680,000 people exposed to computed tomography scans in childhood or adolescence: data linkage study of 11 million Australians,” BMJ, vol. 346, p. f2360, 2013, doi: 10.1136/bmj.f2360.

[6] D. J. Brenner and E. J. Hall, “Computed tomography — an increasing source of radiation exposure,” New England Journal of Medicine, vol. 357, no. 22, pp. 2277–2284, 2007, doi: 10.1056/NEJMra072149.

[7] D. E. Saunders, C. Thompson, R. Gunny, R. Jones, T. Cox, and W. K. Chong, “Magnetic resonance imaging protocols for paediatric neuroradiology,” Pediatric Radiology, vol. 37, no. 8, pp. 789–797, 2007, doi: 10.1007/s00247-007-0462-9.

[8] M. Krönke et al., “Tracked 3D ultrasound and deep neural network-based thyroid segmentation reduce interobserver variability in thyroid volumetry,” PLoS ONE, vol. 17, no. 7, e0268550, 2022, doi: 10.1371/journal.pone.0268550.

[9] G. Alexandridis, E. W. Verschuuren, A. V. Rosendaal, and D. A. Kanhai, “Evidence base for point-of-care ultrasound (POCUS) for the diagnosis of skull fractures in children: a systematic review and meta-analysis,” Emergency Medicine Journal, vol. 39, no. 1, pp. 30–36, 2022, doi: 10.1136/emermed-2020-209893.

[10] C. Caroselli et al., “Accuracy of point-of-care ultrasound in detecting fractures in children: a validation study,” Ultrasound in Medicine and Biology, vol. 47, no. 1, pp. 68–75, 2021, doi: 10.1016/j.ultrasmedbio.2020.09.019.

[11] D. Wood et al., “Ultrasound in forearm fractures: a pragmatic study assessing the utility of point-of-care ultrasound in identifying and managing distal radius fractures,” Emergency Radiology, vol. 28, no. 6, pp. 1107–1112, 2021, doi: 10.1007/s10140-021-01957-8.

[12] C. Mi et al., “Bone disease imaging through the near-infrared-II window,” Nature Communications, vol. 14, p. 6018, 2023, doi: 10.1038/s41467-023-42001-2.

[13] G. Jeffery, R. Fosbury, E. Barrett, C. Hogg, M. R. Carmona, and M. B. Powner, “Longer wavelengths in sunlight pass through the human body and have a systemic impact which improves vision,” Scientific Reports, vol. 15, 09785, 2025, doi: 10.1038/s41598-025-09785-3.

[14] Y. Zhao et al., “Shortwave-infrared meso-patterned imaging enables label-free mapping of tissue water and lipid content,” Nature Communications, vol. 11, p. 5355, 2020, doi: 10.1038/s41467-020-19128-7.

[15] L. V. Wang and H. Wu, Biomedical Optics: Principles and Imaging. Hoboken, NJ, USA: Wiley, 2007.

[16] H. Zhang, D. Salo, D. M. Kim, S. Komarov, Y.-C. Tai, and M. Y. Berezin, “Penetration depth of photons in biological tissues from hyperspectral imaging in shortwave infrared in transmission and reflection geometries,” Journal of Biomedical Optics, vol. 21, no. 12, p. 126006, 2016, doi: 10.1117/1.JBO.21.12.126006.

[17] A. Pifferi et al., “Time-resolved diffuse reflectance using small source–detector separation and fast single-photon gating,” Physical Review Letters, vol. 100, no. 13, p. 138101, 2008, doi: 10.1103/PhysRevLett.100.138101.

[18] T. Vo-Dinh, Ed., Biomedical Photonics Handbook, 2nd ed. Boca Raton, FL, USA: CRC Press, 2014.

[19] D. Xu et al., “In vivo calcium imaging in the near-infrared II window,” Journal of the American Chemical Society, vol. 148, pp. 349–359, 2026, doi: 10.1021/jacs.5c13641.

[20] S. Mironov et al., “Short-wave infrared light imaging measures tissue moisture and distinguishes superficial from deep burns,” Wound Repair and Regeneration, vol. 28, no. 2, pp. 185–193, 2020, doi: 10.1111/wrr.12779.

[21] B. F. Jones, “A reappraisal of the use of infrared thermal image analysis in medicine,” IEEE Transactions on Medical Imaging, vol. 17, no. 6, pp. 1019–1027, 1998, doi: 10.1109/42.746635.

[22] M. R. Stringer et al., “The analysis of human cortical bone by terahertz time-domain spectroscopy,” Physics in Medicine and Biology, vol. 50, no. 14, pp. 3211–3219, 2005, doi: 10.1088/0031-9155/50/14/001.

[23] H. Yan et al., “Handheld near-infrared spectroscopy: state-of-the-art instrumentation and applications in material identification, food authentication, and environmental investigations,” Chemosensors, vol. 11, no. 5, p. 272, 2023, doi: 10.3390/chemosensors11050272.

[24] S. Mukhtar, A. Arbabi, and J. Viegas, “Advances in spectral imaging: a review of techniques and technologies,” IEEE Access, vol. 13, pp. 31678–31703, 2025, doi: 10.1109/ACCESS.2025.3544476.

[25] International Electrotechnical Commission, IEC 60825-1: Safety of Laser Products — Part 1: Equipment Classification and Requirements, Edition 3.0. Geneva, Switzerland: IEC, 2014.

[26] P. Isola, J.-Y. Zhu, T. Zhou, and A. A. Efros, “Image-to-image translation with conditional adversarial networks,” in Proc. IEEE Conf. Computer Vision and Pattern Recognition (CVPR), 2017, pp. 5967–5976, doi: 10.1109/CVPR.2017.632.

[27] J.-Y. Zhu, T. Park, P. Isola, and A. A. Efros, “Unpaired image-to-image translation using cycle-consistent adversarial networks,” in Proc. IEEE Int. Conf. Computer Vision (ICCV), 2017, pp. 2242–2251, doi: 10.1109/ICCV.2017.244.

[28] K. Koshino et al., “Narrative review of generative adversarial networks in medical and molecular imaging,” Annals of Translational Medicine, vol. 9, no. 9, p. 821, 2021, doi: 10.21037/atm-20-6325.

[29] L. Han et al., “All-in-one medical image-to-image translation,” Cell Reports Methods, vol. 5, p. 101138, 2025, doi: 10.1016/j.crmeth.2025.101138.

[30] M. Tahmid, M. S. Alam, N. Rao, and K. M. A. Ashrafi, “Image-to-image translation with conditional adversarial networks,” in Proc. IEEE WIECON-ECE, 2023, doi: 10.1109/WIECON-ECE60392.2023.10456447.

[31] H. Cao et al., “Swin-Unet: Unet-like pure transformer for medical image segmentation,” in Proc. ECCV Workshops, 2022, pp. 205–218.

[32] Z. Liu et al., “Swin Transformer: hierarchical vision transformer using shifted windows,” in Proc. IEEE Int. Conf. Computer

Vision (ICCV), 2021, pp. 10012–10022, doi: 10.1109/ICCV48922.2021.00986.

[33] D. DeTone, T. Malisiewicz, and A. Rabinovich, “SuperPoint: self-supervised interest point detection and description,” in Proc. IEEE CVPR Workshops, 2018, pp. 224–236, doi: 10.1109/CVPRW.2018.00060.

[34] P.-E. Sarlin, D. DeTone, T. Malisiewicz, and A. Rabinovich, “SuperGlue: learning feature matching with graph neural networks,” in Proc. IEEE CVPR, 2020, pp. 4938–4947, doi: 10.1109/CVPR42600.2020.00499.

[35] X. Zhao, X. Wu, W. Chen, L. Chen, Q. Li, and S. Wang, “ALIKED: a lighter keypoint and descriptor extraction network via deformable transformation,” IEEE Transactions on Instrumentation and Measurement, vol. 72, pp. 1–16, 2023, doi: 10.1109/TIM.2023.3271000.

[36] P. Lindenberger, P.-E. Sarlin, and M. Pollefeys, “LightGlue: local feature matching at light speed,” in Proc. IEEE ICCV, 2023, pp. 17627–17638, doi: 10.1109/ICCV51070.2023.01616.

[37] L. Morelli, F. Ioli, F. Maiwald, G. Mazzacca, F. Menna, and F. Remondino, “Deep-image-matching: a toolbox for multiview image matching of complex scenarios,” Int. Arch. Photogramm. Remote Sens. Spat. Inf. Sci., vol. XLVIII-2/W4-2024, pp. 309–316, 2024, doi: 10.5194/isprs-archives-XLVIII-2-W4-2024-309-2024.

[38] H. Chang, A. Koschan, M. Abidi, S. G. Kong, and C. H. Won, “Multispectral visible and infrared imaging for face recognition,” in Proc. IEEE CVPR Workshops, 2008, pp. 1–6, doi: 10.1109/CVPRW.2008.4563054.

[39] N. Parri, B. J. Crosby, C. Glass, et al., “Ability of emergency ultrasonography to detect pediatric skull fractures: a prospective, observational study,” Journal of Emergency Medicine, vol. 44, no. 1, pp. 135–141, 2013, doi: 10.1016/j.jemermed.2012.02.038.

[40] M. Salvi et al., “Multi-modality approaches for medical support systems: a systematic review of the last decade,” Information Fusion, vol. 103, p. 102134, 2024, doi: 10.1016/j.inffus.2023.102134.

[41] M. U. Suleman et al., “Assessing the generalizability of artificial intelligence in radiology: a systematic review of performance across different clinical settings,” Annals of Medicine and Surgery, vol. 87, pp. 8803–8811, 2025, doi: 10.1097/MS9.0000000000004166.

[42] M. W. Sjoding, R. P. Dickson, T. J. Iwashyna, S. E. Gay, and T. S. Valley, “Racial bias in pulse oximetry measurement,” New England Journal of Medicine, vol. 383, no. 25, pp. 2477–2478, 2020, doi: 10.1056/NEJMc2029240.

[43] I. A. Atalay Appak et al., “Multispectral extended depth-of-field fluorescence microscopy with co-designed meta-optics and neural reconstruction,” Light: Science and Applications, vol. 15, p. 242, 2026, doi: 10.1038/s41377-026-02337-y.

[44] International Electrotechnical Commission, IEC 60601-1: Medical Electrical Equipment — Part 1: General Requirements for Basic Safety and Essential Performance, Edition 3.2. Geneva, Switzerland: IEC, 2020.

[45] U.S. Food and Drug Administration, Marketing Submission Recommendations for a Predetermined Change Control Plan for Artificial Intelligence/Machine Learning-Enabled Device Software Functions: Guidance for Industry and Food and Drug Administration Staff. Silver Spring, MD, USA: FDA, 2024.

[46] L. Seyyed-Kalantari, H. Zhang, M. B. A. McDermott, I. Y. Chen, and M. Ghassemi, “Underdiagnosis bias of artificial intelligence algorithms applied to chest radiographs in under-served patient populations,” Nature Medicine, vol. 27, no. 12, pp. 2176–2182, 2021, doi: 10.1038/s41591-021-01595-0.

[47] T. Gao, T. Fan, W. Leng, S. Yu, and Q. Lyu, “Artificial intelligence in acute and critical care: current challenges and strategic solutions,” Frontiers in Public Health, vol. 14, p. 1818726, 2026, doi: 10.3389/fpubh.2026.1818726.

[48] World Medical Association, “Declaration of Helsinki — ethical principles for medical research involving human subjects,” JAMA, vol. 310, no. 20, pp. 2191–2194, 2013, doi: 10.1001/jama.2013.281053.

[49] J. Napp, J. E. Mathejczyk, and F. Alves, “Optical imaging in vivo with a focus on paediatric disease: technical progress, current preclinical and clinical applications and future perspectives,” Pediatric Radiology, vol. 41, no. 2, pp. 161–175, 2011, doi: 10.1007/s00247-010-1907-0.

[50] E. Ruggiero, S. Alonso-de Castro, A. Habtemariam, and L. Salassa, “Upconverting nanoparticles for the near infrared photoactivation of transition metal complexes: new opportunities and challenges in medicinal inorganic photochemistry,” Dalton Transactions, vol. 45, pp. 13012–13020, 2016, doi: 10.1039/c6dt01428c.

[51] R. Tao, J. Gröhl, L. Hacker, A. Pifferi, D. Roblyer, and S. E. Bohndiek, “Tutorial on methods for estimation of optical absorption and scattering properties of tissue,” Journal of Biomedical Optics, vol. 29, no. 6, p. 060801, 2024, doi: 10.1117/1.JBO.29.6.060801.

[52] A. Y. Denault, M. Shaaban-Ali, A. Cournoyer, A. Benkreira, and T. Mailhot, “Near-infrared spectroscopy,” in Neuromonitoring Techniques. Amsterdam, The Netherlands: Elsevier, 2018, ch. 7, pp. 179–233, doi: 10.1016/B978-0-12-809915-5.00007-3.

[53] J. Mongan, L. Moy, and C. E. Kahn, Jr., “Checklist for Artificial Intelligence in Medical Imaging (CLAIM): a guide for authors and reviewers,” Radiology: Artificial Intelligence, vol. 2, no. 2, p. e200029, 2020, doi: 10.1148/ryai.2020200029.

[54] A. Krumholz, D. M. Shcherbakova, J. Xia, L. V. Wang, and V. V. Verkhusha, “Multicontrast photoacoustic in vivo imaging using near-infrared fluorescent proteins,” Scientific Reports, vol. 4, p. 3939, 2014, doi: 10.1038/srep03939.

[55] F. F. Jöbsis, “Noninvasive, infrared monitoring of cerebral and myocardial oxygen sufficiency and circulatory parameters,” Science, vol. 198, no. 4323, pp. 1264–1267, 1977, doi: 10.1126/science.929199.

[56] M. Ferrari and V. Quaresima, “A brief review on the history of human functional near-infrared spectroscopy (fNIRS) development and fields of application,” NeuroImage, vol. 63, no. 2, pp. 921–935, 2012, doi: 10.1016/j.neuroimage.2012.03.049.

[57] Y. Huang, X. Zhan, H. Zeng, S. Li, J. Shi, Z. Cui, Q. Fan, B. Li, Y. Sui, F. Liang, and Z. Song, “A systematic review of functional near-infrared spectroscopy-based task paradigms in stroke rehabilitation,” Frontiers in Human Neuroscience, vol. 19, p. 1633142, 2026, doi: 10.3389/fnhum.2025.1633142.

[58] W. L. Chen, J. Wagner, N. Heugel, J. Sugar, Y. W. Lee, L. Conant, et al., “Functional near-infrared spectroscopy and its clinical application in the field of neuroscience: advances and future directions,” Frontiers in Neuroscience, vol. 14, p. 724, 2020, doi: 10.3389/fnins.2020.00724.

[59] M. Wolf, M. Ferrari, and V. Quaresima, “Progress of near-infrared spectroscopy and topography for brain and muscle clinical applications,” Journal of Biomedical Optics, vol. 12, no. 6, p. 062104, 2007, doi: 10.1117/1.2804899.

[60] D. Nizamutdinov, C. Ezeudu, E. Wu, J. H. Huang, and S. S. Yi, “Transcranial near-infrared light in treatment of neurodegenerative diseases,” Frontiers in Pharmacology, vol. 13, p. 965788, 2022, doi: 10.3389/fphar.2022.965788.

[61] T. Wang, D. G. Ouzounov, C. Wu, N. G. Horton, B. Zhang, C.-H. Wu, et al., “Three-photon imaging of mouse brain

structure and function through the intact skull," Nature Methods, vol. 15, no. 10, pp. 789–792, 2018, doi: 10.1038/s41592-018-0115-y.

[62] A. Pitzschke, B. Lovisa, O. Seydoux, M. Zellweger, M. Pfleiderer, Y. Tardy, et al., "Red and NIR light dosimetry in the human deep brain," Physics in Medicine and Biology, vol. 60, no. 7, pp. 2921–2937, 2015, doi: 10.1088/0031-9155/60/7/2921.

[63] N. E. Pretorius, N. Anderson, A. S. Forrest, and K. B. Walsh, "Near infrared spectroscopy for the characterisation of bovine teeth in terms of sex, diet, tooth type and place of origin," Journal of Near Infrared Spectroscopy, vol. 32, no. 6, pp. 207–217, 2024, doi: 10.1177/09670335241300641.